\documentclass{article}
\usepackage{spconf,amsmath,graphicx}\usepackage{float}
\usepackage{lipsum}
\usepackage{graphicx}

\usepackage{enumitem}
\setlist{nosep, leftmargin=14pt}

\usepackage{mwe} 
\usepackage{amsfonts} 

\title{Edge-preserving Domain Adaptation for semantic segmentation of Medical Images}
\name{Thong Vo, Naimul Khan}
 \address{Ryerson University}
\begin{document}
%
\maketitle
\begin{abstract}
Domain Adaptation is a technique to address the lack of massive amounts of labeled data in unseen environments. Unsupervised domain adaptation is proposed to adapt a model to new modalities using solely labeled source data and unlabeled target domain data. Though many image-spaces domain adaptation methods have been proposed to capture pixel-level domain-shift, such techniques may fail to maintain high-level semantic information for the segmentation task. For the case of biomedical images, fine details such as blood vessels can be lost during the image transformation operations between domains.  In this work, we propose a model that adapts between domains using cycle-consistent loss while maintaining edge details of the original images by enforcing an edge-based loss during the adaptation process. We demonstrate the effectiveness of our algorithm by comparing it to other approaches on two eye fundus vessels segmentation datasets. We achieve 1.1 to 9.2 increment in DICE score compared to the SOTA and $\sim$5.2 increments compared to a vanilla CycleGAN implementation.
\end{abstract}
\begin{keywords}
Generative Adversarial Networks, Domain
Adaptation, Edge Detector
\end{keywords}
\section{Introduction}
\label{sec:intro}

CNNs have demonstrated their superiority in various semantic segmentation applications such as medical image analysis \cite{cnnsegment}, object detection, autonomous driving \cite{long2015fully}. The success of CNNs, however, is limited to supervised training using the availability of huge manually annotated datatsets. In the context of medical images, target objects are often small and densely packed making the annotation process laborious. Additionally, unlike general objects datasets that can be annotated effectively by any individual with minimal guidance, medical image segmentation often requires trained experts with experience in specific domains. This makes it challenging to obtain annotated medical image datasets.

The variation between the training data (source domain) and testing data (target domain), or domain shift, is a major area of concern. In the context of biomedical applications, domain shift and dataset bias could be due to various reasons such as image acquisition methods, illumination variations, imaging resolution, or even more fundamental differences like variations in the blood vessels that are imaged. Domain shift can be alleviated by fine-tuning a model \cite{brainmri}, previously trained on a source dataset, with samples withdraw from a target dataset. Such methods, though, still require some annotated images in the target dataset to calibrate the trained model. Another approach used to solve this problem is domain adaptation. Domain adaptation techniques usually attempt to align the source and target data distributions in the image spaces \cite{Bousmalis2017UnsupervisedPD} or extracted feature spaces\cite{Ganin2016DomainAdversarialTO}.

\begin{figure}[t!]
    \centering
    \includegraphics[width=0.47\textwidth]{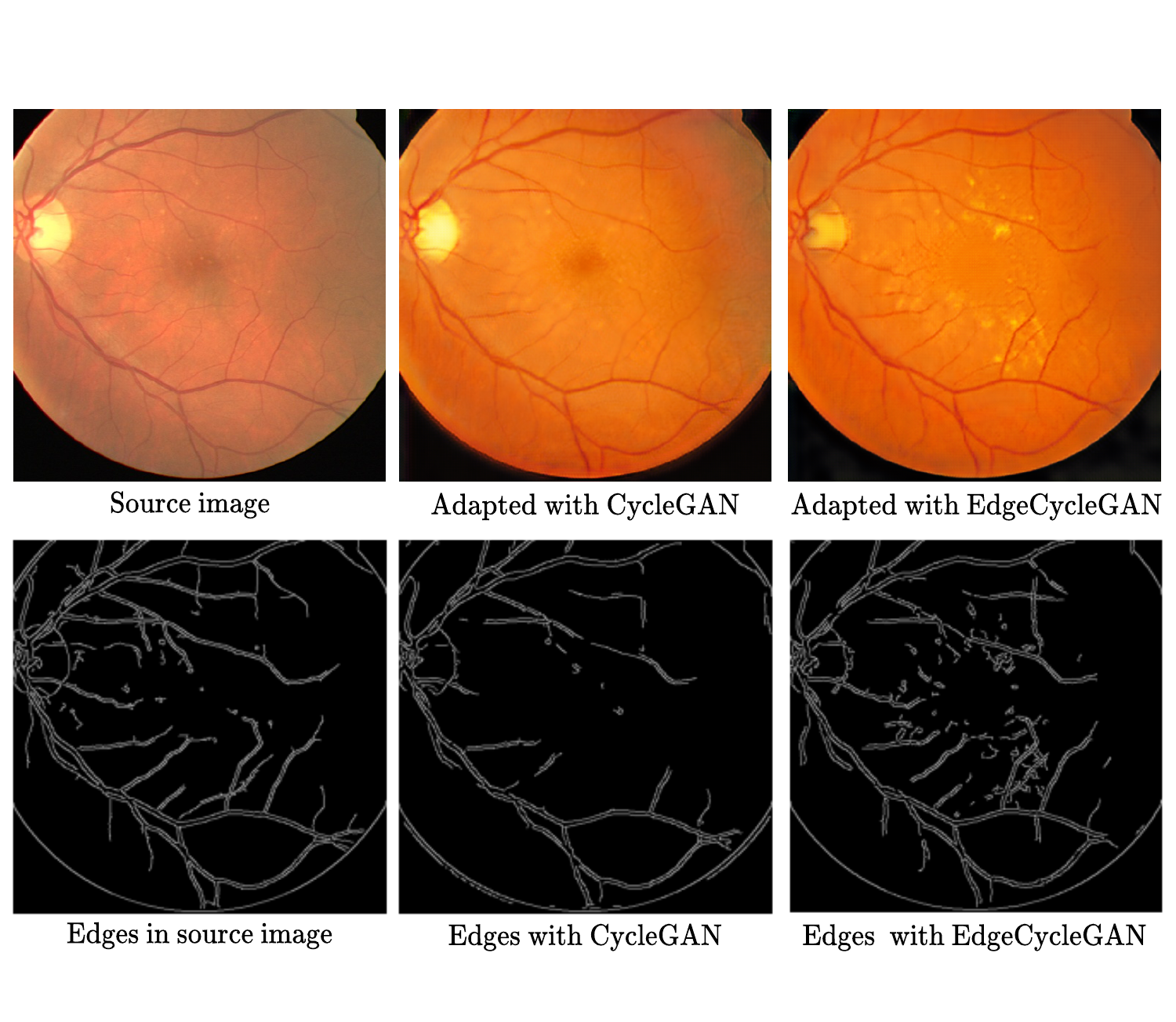}
    \caption{Original image and adapted images comparison. The extracted edges in the lower half demonstrate the edge details in each picture. The EdgeCycleGAN preserves border information in the original image, which leads to a better segmentation result.}
    \label{fig:mesh1}
\end{figure}

In this paper, we propose an edge-preserving domain adaptation on top of a segmentor network, which adapts representations at the pixel level while enforcing semantic consistency and edge details preservation. The domain adaptation leverages the cycle-consistency loss\cite{cyclegan} that enforces a consistent image translation between the source and the target domain. Meanwhile, an edge-based loss term is added to protect the edge information throughout the translation process. We validate our algorithm on two medical image segmentation datasets, and observe comparable or better performance to SOTA methods\cite{javanmardi2018domain} based on adversarial training.

\section{RELATED WORK}
\label{sec:relatedwork}
\subsection{Image Segmentation}
CNNs have excellent feature extraction capabilities for visual tasks, and the models do not require manual extractions of image features or excessive preprocessing steps. Therefore, CNNs have been employed in image segmentation in different sets of architectures. One breakthrough design is the Fully Convolutional Networks (FCN)\cite{long2015fully}. FCNs take variable input sizes and output related segmentation predictions for every pixel of that input. In this setting, skip connections are used to skip features from the contracting path to the expanding path in order to recover spatial information lost along the downsampling path. As a follow-up, the U-Net\cite{ronneberger2015u} improved upon FCNs architecture by having a larger number of features in the upsampling path and adding more convolution layers. In this way, U-net has shown significant improvements from FCNs architecture, especially for the medical image domain[cite]. The reason why U-Net is suitable for medical image segmentation is that its structure can simultaneously combine low-level and high-level information. The high-level information contains semantic information, while the low-level ones affect the accuracy of the model.   While there are various incremental improvement proposals for the U-Net\cite{Zhou2018UNetAN}, the original architecture is still a strong candidate for segmentation tasks because of its effectiveness and uncomplicated design.

\subsection{Domain Adaptation}
Various domain adaptation approaches \cite{Long2015LearningTF} have been explored in both semi-supervised and unsupervised settings. Semi-supervised domain adaptation tries to exploit the knowledge from the source domain and leverage a certain number of unlabeled examples and a few labeled ones from the target domain to learn a target model. The constraint of this approach is the manual data annotation requirement for a small number of images which might be difficult to obtain in the medical setting. Meanwhile, the unsupervised domain adaptation is more appealing because it does not require additional labels for the target domain. Researchers have been trying to tackle this challenging problem from different angles.

One line of research aims to learn domain-agnostic features across domains via a domain classifier loss. Most of the models adopt a siamese architecture proposed by the Domain Adversarial Neural Network (DANN) \cite{Ganin2016DomainAdversarialTO} that employs a gradient reversal layer to guide the domain invariant feature learning. In this way, a DANN-based model for eye vasculature segmentation was proposed\cite{javanmardi2018domain}. A U-Net and a domain discriminator are co-trained to generate a domain-agnostic image segmentation model. The authors consider the whole output probability map for domain classification instead of the feature map as suggested in \cite{Hoffman2018CyCADACA}. Experiments on the DRIVE\cite{drive} and STARE\cite{stare} datasets demonstrated its effectiveness in the domain adaptation task. To enhance the model's attention edges, edge detectors are added to improve the performance of the adversarial learning process \cite{Yan2019EdgeGuidedOA}. Assessments with images from three independent Magnetic Resonance Imaging vendors (Philips, Siemens, and GE) show the benefits of this addition.

Another line of research focuses on the image-level alignment by leveraging the Generative Adversarial Network (GAN) \cite{Goodfellow2014GenerativeAN} to boost the domain adaptation results. CycleGAN model \cite{Zhu2017UnpairedIT} has been proposed to translate one image domain into another without the presence of aligned image pairs. This approach introduces a cycle consistency loss to encourage information preservation while translating back and forward between domains by measuring the difference between the input image and the reconstructed image. For example, CycleGAN can translate images between the MR image and CT image \cite{Wolterink2017DeepMT}. Another example is the Noise Adaptation Generative Adversarial Network (NAGAN)\cite{Zhang2020NoiseAG} which improves the eye blood vessel segmentation by formulating the noise adaptation task as an image-to-image translation task. In recent work, the CyCADA model\cite{Hoffman2018CyCADACA} incorporates both the feature level adaptations and image-level adaptations to further improve the performance of CycleGAN based domain adaptation. This work is different from our approach in the sense that we introduce an additional edge-based loss that encourages the model to preserve edge information when translating between domains. 

\begin{figure*}[ht]
\centering
\includegraphics[width=1\textwidth]{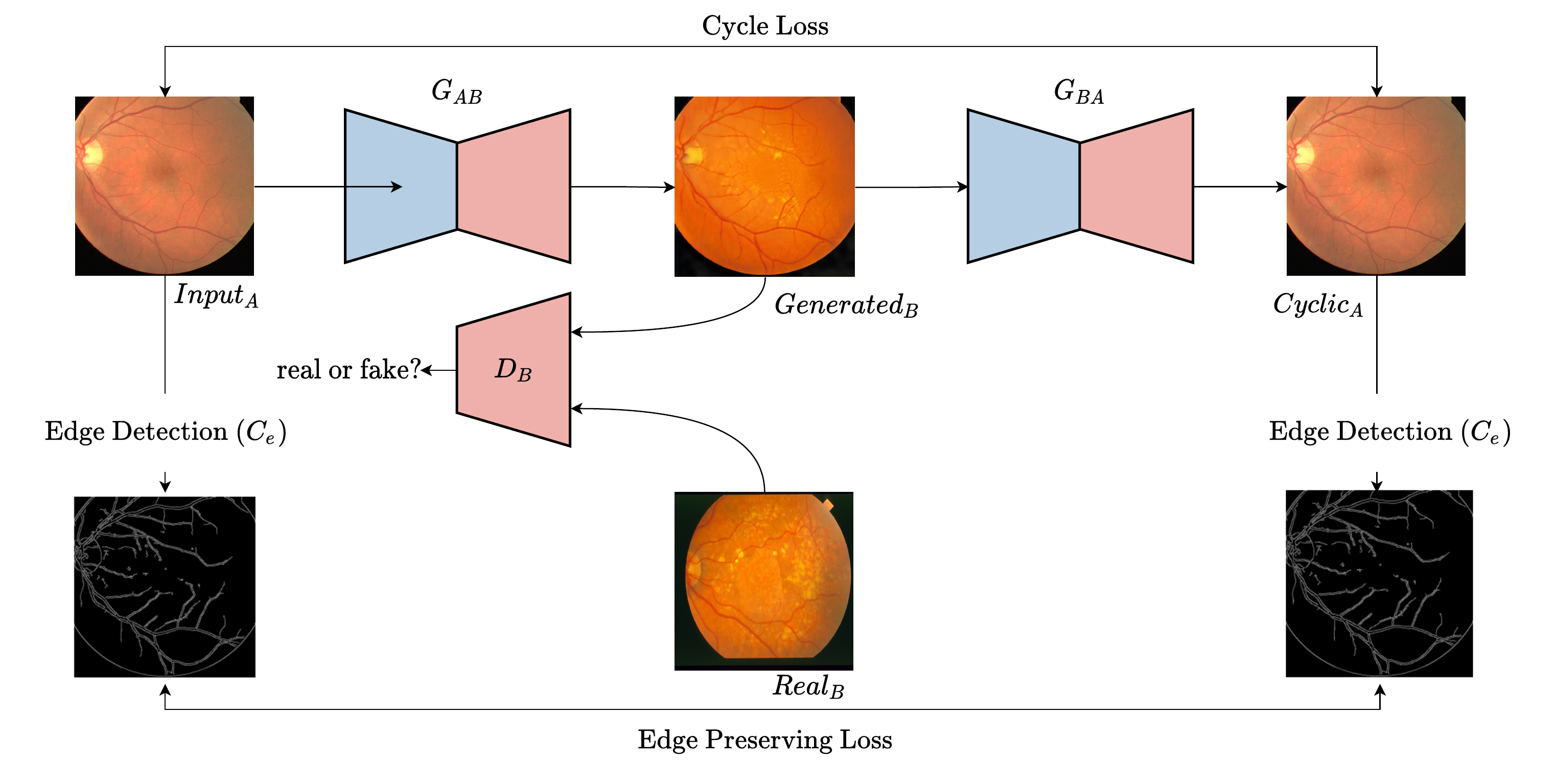}
\caption{Edge-preserving CycleGAN architecture. By adding the edge-preserving loss on top of the CycleGAN, we encourage the model to protect bordering information during domain mapping.}
\end{figure*}

\section{Approach}
\label{sec:approach}
\subsection{Problem setting}
Our proposed method consists of two networks, one for the image adaptation and one for the segmentation task. We first describe the EdgeCycleGAN as the adaptation method, then followed by a U-Net network to perform the segmentation task.

\subsection{EdgeCycleGAN}
The proposed EdgeCycleGAN comprises the original CycleGAN configurations and an additional edge-based loss to preserve the targeted edge information. CycleGAN has two paired generator-discriminator modules, which are capable of learning two mappings, i.e., from domain $\mathrm{A}$ to domain $\mathrm{B}\left\{G, D_{B}\right\}$ and the inverse $\mathrm{B}$ to A $\left\{F, D_{A}\right\}$. The generators $\left(G, F\right)$ translate images between the source and target domains, while the discriminators $\left(D_{A}, D_{B}\right)$ aim to distinguish the original data from the translated ones. Thereby, the generators and discriminators are gradually updated during this adversarial competition.
The original CycleGAN is supervised by two losses, the adversarial loss $\mathcal{L}_{a d v}$ and the cycle-consistency loss $\mathcal{L}_{\text {cyc }}$. The adversarial loss encourages local realism of the translated data. Taking the translation from domain A to domain B as an example, the adversarial loss can be written as:
$$
\begin{aligned}
\mathcal{L}_{\mathrm{GAN}}\left(G, D_{B}, A, B\right) &=\mathbb{E}_{y \sim p_{\text {data }}(y)}\left[\log D_{B}(y)\right] \\
&+\mathbb{E}_{x \sim p_{\text {data }}(x)}\left[\log \left(1-D_{B}(G(x))\right]\right.
\end{aligned}
$$
where $p_{data(x)}$ and $p_{data(y)}$ represent the sample distributions of domain A and B, respectively; $x$ and $y$ are samples from domain $\mathrm{A}$ and $\mathrm{B}$, respectively.

The cycle-consistency loss $\mathcal{L}_{c y c}$ relieves the requirement of paired training data. The concept of cycle-consistency loss is that the converted data from the target domain can be translated back to the source domain, which can be denoted as:
$$
\begin{aligned}
\mathcal{L}_{\mathrm{cyc}}(G, F) &=\mathbb{E}_{x \sim p_{\text {data }}(x)}\left[\|F(G(x))-x\|_{1}\right] \\
&+\mathbb{E}_{y \sim p_{\text {data }}(y)}\left[\|G(F(y))-y\|_{1}\right]
\end{aligned}
$$
In this paper, we propose an additional loss term to preserve the edge information. The Canny edge extractor\cite{canny1986computational} $C_{e}$ highlights edge details from the input image tensors. The loss term $\mathcal{L}_{\mathrm{edge}}$ penalizes the model when edge details are lost during the image transformations. The edges of reconstructed images are compared with the edges of the original images.
$$
\begin{aligned}
\mathcal{L}_{
\mathrm{edge}}(G, F) &=
\mathbb{E}_{x 
\sim p_{\text {data }}(x)}\left[\|C_{e}(F(G(x)))-C_{e}(x)\|_{1}\right] \\
&+\mathbb{E}_{y \sim p_{\text {data }}(y)}\left[\|C_{e}(G(F(y)))-C_{e}(y)\|_{1}\right]
\end{aligned}
$$
Combining the three loss terms, the full objective is:
$$
\begin{aligned}
\mathcal{L}\left(G, F, D_{A}, D_{B}\right) &=\mathcal{L}_{\mathrm{GAN}}\left(G, D_{B}, A, B\right) \\
&+\mathcal{L}_{\mathrm{GAN}}\left(F, D_{A}, B, A\right) \\
&+\lambda \mathcal{L}_{\mathrm{cyc}}(G, F) \\
&+\gamma \mathcal{L}_{\mathrm{edge}}(G, F) 
\end{aligned}
$$
where $\lambda$ and $\gamma$ control the relative importance of the objectives. EdgeCycleGAN aim to solve:
$$
G^{*}, F^{*}=\arg \min _{G, F} \max _{D_{A}, D_{B}} \mathcal{L}\left(C_e,G, F, D_{A}, D_{B}\right)
$$
The trained generator pairs ${G^{*},F^{*}}$ are employed to translate images from the source domain to the target domain in the following segmentation operation.

\subsection{Segmentation network}
This network follows a configuration of a U-Net supervised segmentation model and is trained with generated images. Let source domain images be $X_{s}$ with pixel-wise labels $Y_{s}$ and let the target domain images be $X_{t}.$. Per problem setting, the target domain label $Y_t$ is used only for testing purposes. We generate a source image in the target domain by feeding $G_{s \rightarrow t}(X_{s})$ to the segmentor $U$ which we call $U(G_{s \rightarrow t}(X_{s}))$. The generator network $G$ is freeze during the optimization of the segmentation network. The outputs from the U-Net are fed into the softmax loss layer with their corresponding label ground truth. The final loss to optimize for the network is 
$\mathcal{L}_{\mathrm{seg}} \left(U\left(G_{s \rightarrow t}(X_{s}\right)), Y_{s}\right)$.
The intuition behind the proposed architecture is that if the model is trained on the images that have been transformed into the target domain, it would generalize better when tested in the target domain. Because the images are originated in the source domains, their related labels $Y_{s}$ are available for training. Besides, we had tried to add domain discriminators to encourage the segmentation network to learn domain-invariant features \cite{Hoffman2018CyCADACA}, though we found out that such discriminator losses were mostly stable because the transformed images are visually similar to the real images in the target domain.

\section{EXPERIMENTS}
\label{sec:experiments}
We validate our method using two popular eye vessel extraction datasets \cite{drive}\cite{stare}. The two datasets alternate their roles as the source and target datasets in two domain adaptation experiments. The DRIVE (Digital Retinal Images for Vessel Extraction) dataset \cite{drive} includes a total of 40 annotated color fundus images that are divided into two sets. The training set and testing set both containing 20 images. The STARE (STructured Analysis of the Retina) dataset\cite{stare} includes 20 images with corresponding labels.

The first experiment involved training the segmentation model without any adaptation method. The purpose of this step is to set a baseline for the segmentation model. The segmentation network is train with the source dataset $(X_s,Y_s)$ and the trained model is tested with the target domain dataset $(X_t,Y_t)$. We perform this experiment two times by alternating the roles of the DRIVE and STARE datasets. 

In the second experiment, we first train the EdgeCycleGAN model for two domains with DRIVE dataset as domain $A$ and STARE dataset as domain $B$. As the number of sample in the two datasets are limited, we introduce minor rotation, shift augmentations to increase the data samples for the GAN model. All images are scaled to the $512\times512$ sized patches, and only images are considered for training, all related labels are ignored in this step. We generated 400 patches for both the source and target domain. The EdgeCycleGAN is trained for 200 epochs with $\lambda = 10$ and $\gamma = 3$. The $\lambda$ value follows the suggestion from the original CycleGAN\cite{cyclegan} implementation while the $\gamma$ value was obtained from hyperparameter tuning. In addition, the $\gamma$ is set to be smaller than the $\lambda$ value so the edge information will not dominate over other information. Following that, we use one trained generator to generate DRIVE images in the STARE domain, then train the segmentation model with the generated images and the original segmentation labels in DRIVE dataset. The adapted segmentation model is tested with the STARE images and corresponding labels. Similarly, we used the other generator to generate images in the DRIVE domain and test the adapted model with DRIVE dataset.

For the final experiment, we perform similar steps as the second experiment with the original CycleGAN model instead of our proposed EdgeCycleGAN architectures. Our results for every experiment are reported in Table \ref{tab:table_evaluation}.

\begin{table}[]
\begin{tabular}{c|cc}
\hline
Model        & \small DRIVE $\rightarrow$ \text{STARE} & \small STARE $\rightarrow$ \text{DRIVE} \\ \hline
Baseline     & 68.08                     & 63.82                    \\
Javanmardi \textit{et al.}\cite{javanmardi2018domain}   & 76.75                     & 67.09                     \\
CycleGAN     & 72.82                     & 71.05                       \\
EdgeCycleGAN & \textbf{77.84}                     & \textbf{76.29}                    
\end{tabular}
\caption{\label{tab:table_evaluation}The f-scores for segmentation results are evaluated for source $\rightarrow$ target adaptation. Our model is competitive with or outperforms state-of-the-art models for each task.}
\end{table}

According to Table \ref{tab:table_evaluation}, the proposed method demonstrates considerable improvements for the unsupervised domain adaptation task. In the DRIVE $\rightarrow$ STARE transfer, the model f-score increases from 68.08 baseline result to 77.84, exceeding the 76.75 value achieved by \cite{javanmardi2018domain}. In the remaining task, STARE $\rightarrow$ DRIVE, the EdgeCycleGAN yields
a significant boost from a 63.82 baseline to 76.29. Also, by adding the edge-preserving loss, the model gains an average of 5 points over the vanilla CycleGAN adaptation model. We also observe that this method provides comparable increments for both adaptation directions as the DICE score for DRIVE $\rightarrow$ STARE and STARE $\rightarrow$ DRIVE are close while some other adaptations methods will favor one direction to the other.
\section{CONCLUSIONS}
\label{sec:conclusions}
In this paper, we demonstrated a novel Edge-preserving Domain Adaptation for semantic segmentation of medical images. We introduced an additional edge-based loss on top of the original CycleGAN model to preserve boundary information when translating images between domains. We also exhibit the improvements of F1 score in target domain testing  against other methods, and the high quality transferred image in the target domain.


\bibliographystyle{IEEEbib}
\bibliography{strings,refs}

\begin{thebibliography}{10}

\bibitem{cnnsegment}
Fausto Milletari, Nassir Navab, and Seyed-Ahmad Ahmadi,
\newblock ``V-net: Fully convolutional neural networks for volumetric medical
  image segmentation,''
\newblock in {\em 2016 Fourth International Conference on 3D Vision (3DV)},
  2016, pp. 565--571.

\bibitem{long2015fully}
Jonathan Long, Evan Shelhamer, and Trevor Darrell,
\newblock ``Fully convolutional networks for semantic segmentation,''
\newblock in {\em Proceedings of the IEEE conference on computer vision and
  pattern recognition}, 2015, pp. 3431--3440.

\bibitem{brainmri}
Mohsen Ghafoorian, Alireza Mehrtash, Tina Kapur, Nico Karssemeijer, Elena
  Marchiori, Mehran Pesteie, Charles R.~G. Guttmann, F.‐E. Leeuw, Clare~M.
  Tempany, Bram van Ginneken, Andriy~Y Fedorov, Purang Abolmaesumi, Bram
  Platel, and William~M. Wells,
\newblock ``Transfer learning for domain adaptation in mri: Application in
  brain lesion segmentation,''
\newblock in {\em MICCAI}, 2017.

\bibitem{Bousmalis2017UnsupervisedPD}
Konstantinos Bousmalis, Nathan Silberman, David Dohan, Dumitru Erhan, and Dilip
  Krishnan,
\newblock ``Unsupervised pixel-level domain adaptation with generative
  adversarial networks,''
\newblock {\em 2017 IEEE Conference on Computer Vision and Pattern Recognition
  (CVPR)}, pp. 95--104, 2017.

\bibitem{Ganin2016DomainAdversarialTO}
Yaroslav Ganin, E.~Ustinova, Hana Ajakan, Pascal Germain, H.~Larochelle,
  François Laviolette, Mario Marchand, and Victor~S. Lempitsky,
\newblock ``Domain-adversarial training of neural networks,''
\newblock {\em J. Mach. Learn. Res.}, vol. 17, pp. 59:1--59:35, 2016.

\bibitem{cyclegan}
Jun-Yan Zhu, Taesung Park, Phillip Isola, and Alexei~A Efros,
\newblock ``Unpaired image-to-image translation using cycle-consistent
  adversarial networks,''
\newblock in {\em Proceedings of the IEEE international conference on computer
  vision}, 2017, pp. 2223--2232.

\bibitem{javanmardi2018domain}
Mehran Javanmardi and Tolga Tasdizen,
\newblock ``Domain adaptation for biomedical image segmentation using
  adversarial training,''
\newblock in {\em 2018 IEEE 15th International Symposium on Biomedical Imaging
  (ISBI 2018)}. IEEE, 2018, pp. 554--558.

\bibitem{ronneberger2015u}
Olaf Ronneberger, Philipp Fischer, and Thomas Brox,
\newblock ``U-net: Convolutional networks for biomedical image segmentation,''
\newblock in {\em International Conference on Medical image computing and
  computer-assisted intervention}. Springer, 2015, pp. 234--241.

\bibitem{Zhou2018UNetAN}
Zongwei Zhou, Md~Mahfuzur~Rahman Siddiquee, Nima Tajbakhsh, and Jianming Liang,
\newblock ``Unet++: A nested u-net architecture for medical image
  segmentation,''
\newblock {\em Deep Learning in Medical Image Analysis and Multimodal Learning
  for Clinical Decision Support : 4th International Workshop, DLMIA 2018}, vol.
  11045, pp. 3--11, 2018.

\bibitem{Long2015LearningTF}
Mingsheng Long, Yue Cao, Jianmin Wang, and Michael~I. Jordan,
\newblock ``Learning transferable features with deep adaptation networks,''
\newblock {\em ArXiv}, vol. abs/1502.02791, 2015.

\bibitem{Hoffman2018CyCADACA}
Judy Hoffman, Eric Tzeng, Taesung Park, Jun-Yan Zhu, Phillip Isola, Kate
  Saenko, Alexei~A. Efros, and Trevor Darrell,
\newblock ``Cycada: Cycle-consistent adversarial domain adaptation,''
\newblock in {\em ICML}, 2018.

\bibitem{drive}
Joes Staal, Michael~David Abr{\`a}moff, Meindert Niemeijer, Max~A. Viergever,
  and Bram van Ginneken,
\newblock ``Ridge-based vessel segmentation in color images of the retina,''
\newblock {\em IEEE Transactions on Medical Imaging}, vol. 23, pp. 501--509,
  2004.

\bibitem{stare}
Adam~W. Hoover, Valentina~L. Kouznetsova, and Michael~H. Goldbaum,
\newblock ``Locating blood vessels in retinal images by piecewise threshold
  probing of a matched filter response,''
\newblock {\em IEEE Transactions on Medical Imaging}, vol. 19, pp. 203--210,
  2000.

\bibitem{Yan2019EdgeGuidedOA}
Wenjun Yan, Yuanyuan Wang, Menghua Xia, and Qian Tao,
\newblock ``Edge-guided output adaptor: Highly efficient adaptation module for
  cross-vendor medical image segmentation,''
\newblock {\em IEEE Signal Processing Letters}, vol. 26, pp. 1593--1597, 2019.

\bibitem{Goodfellow2014GenerativeAN}
Ian~J. Goodfellow, Jean Pouget-Abadie, Mehdi Mirza, Bing Xu, David
  Warde-Farley, Sherjil Ozair, Aaron~C. Courville, and Yoshua Bengio,
\newblock ``Generative adversarial nets,''
\newblock in {\em NIPS}, 2014.

\bibitem{Zhu2017UnpairedIT}
Jun-Yan Zhu, Taesung Park, Phillip Isola, and Alexei~A. Efros,
\newblock ``Unpaired image-to-image translation using cycle-consistent
  adversarial networks,''
\newblock {\em 2017 IEEE International Conference on Computer Vision (ICCV)},
  pp. 2242--2251, 2017.

\bibitem{Wolterink2017DeepMT}
Jelmer~M. Wolterink, Anna~M. Dinkla, Mark Savenije, Peter~R. Seevinck, Cornelis
  A.~T. van~den Berg, and Ivana Isgum,
\newblock ``Deep mr to ct synthesis using unpaired data,''
\newblock in {\em SASHIMI@MICCAI}, 2017.

\bibitem{Zhang2020NoiseAG}
Tianyang Zhang, Jun Cheng, H.~Fu, Zaiwang Gu, Yuting Xiao, Kang Zhou, Shenghua
  Gao, Ru~Zheng, and Jiang Liu,
\newblock ``Noise adaptation generative adversarial network for medical image
  analysis,''
\newblock {\em IEEE Transactions on Medical Imaging}, vol. 39, pp. 1149--1159,
  2020.

\bibitem{canny1986computational}
John Canny,
\newblock ``A computational approach to edge detection,''
\newblock {\em IEEE Transactions on pattern analysis and machine intelligence},
  , no. 6, pp. 679--698, 1986.

\end{thebibliography}

\end{document}